\documentclass[10pt,twocolumn,letterpaper]{article}

\usepackage{cvpr}
\usepackage{color}
\usepackage{utfsym}
\usepackage{bbding}
\usepackage{amssymb}
\usepackage{colortbl}
\definecolor{mygray}{gray}{.9}
\usepackage{graphicx}
\usepackage{multirow}
\usepackage{longtable}
\usepackage{subcaption}
\newcounter{magicrownumbers}

\newcommand \footnoteONLYtext[1]
{
    \let \mybackup \thefootnote
    \let \thefootnote \relax
    \footnotetext{#1}
    \let \thefootnote \mybackup
    \let \mybackup \imareallyundefinedcommand
}

\definecolor{cvprblue}{rgb}{0.21,0.49,0.74}
\usepackage[pagebackref,breaklinks,colorlinks,allcolors=cvprblue]{hyperref}

\title{
FaceBench: A Multi-View  Multi-Level Facial Attribute VQA Dataset for Benchmarking Face Perception MLLMs
}

\author{
Xiaoqin Wang$^{1,2,3}$,~
Xusen Ma$^{1,2,3}$,~
Xianxu Hou$^4$,~
Meidan Ding$^{1,2,3}$,~
Yudong Li$^5$,\\
Junliang Chen$^6$,~
Wenting Chen$^7$,~
Xiaoyang Peng$^8$,~
Linlin Shen$^{1,2,3,}$\thanks{Corresponding Author}~\\
\textsuperscript{\rm1}Computer Vision Institute, College of Computer Science and Software Engineering, Shenzhen University\\
\textsuperscript{\rm2}National Engineering Laboratory for Big Data System Computing Technology, Shenzhen University\\
\textsuperscript{\rm3}Guangdong Provincial Key Laboratory of Intelligent Information Processing\\
\textsuperscript{\rm4}AIAC, Xi'an Jiaotong-Liverpool University;~
\textsuperscript{\rm5}Tsinghua University;\\
\textsuperscript{\rm6}The Hong Kong Polytechnic University;~
\textsuperscript{\rm7}City University of Hong Kong;~
\textsuperscript{\rm8}Sun Yat-sen University~\\
{\tt\small wangxiaoqin2022@email.szu.edu.cn;~llshen@szu.edu.cn}
}

\begin{document}

\twocolumn[{
\renewcommand\twocolumn[1][]{#1}
\maketitle
\begin{center}
    \vspace{-21pt}
    \centering
    \captionsetup{type=figure}
    \includegraphics[width=1.0\linewidth]{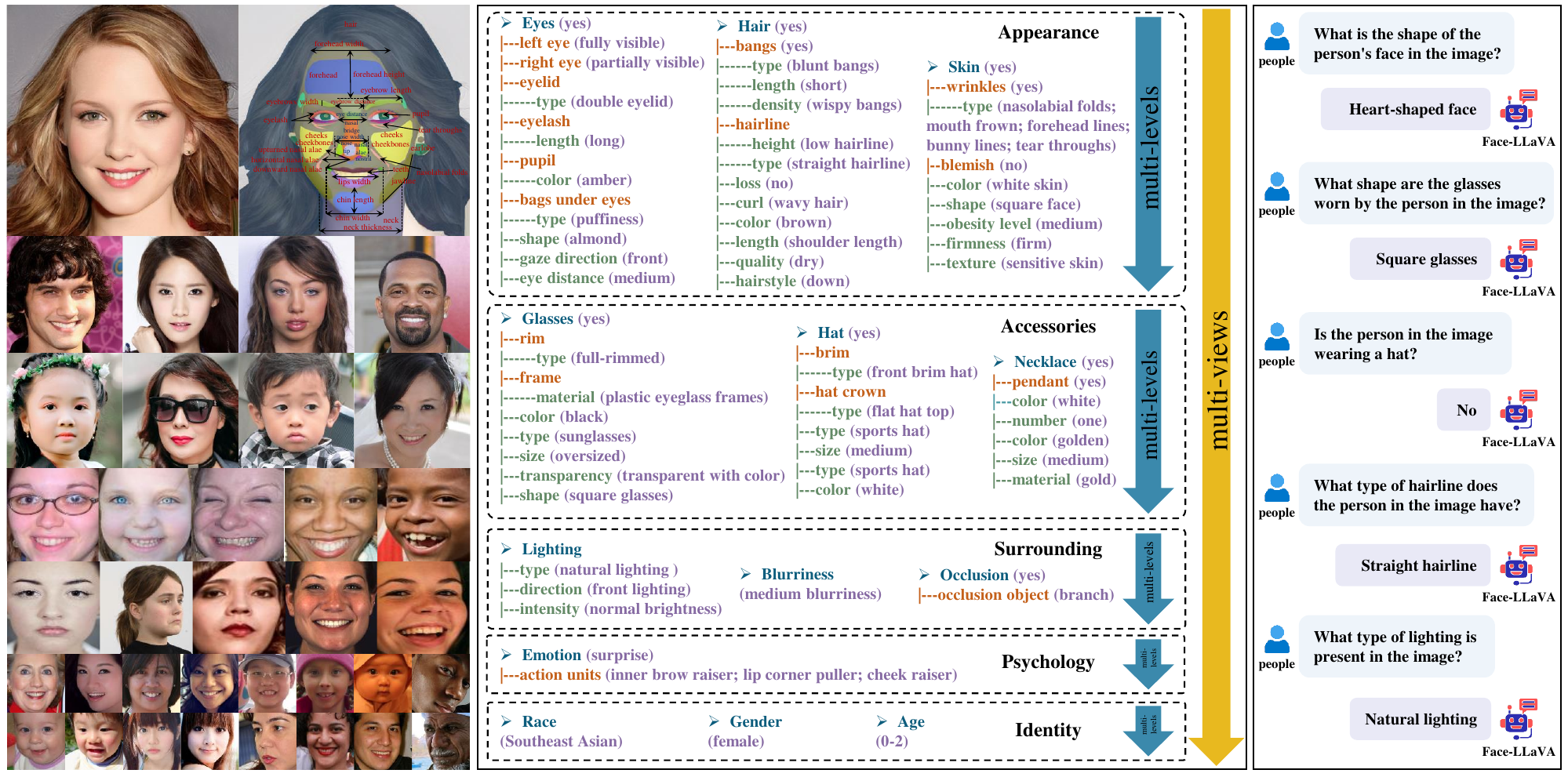}
    \captionof{figure}{Overview of FaceBench. \textbf{Left}: Example of face images, including a face region mask. \textbf{Center}: FaceBench covers multi-views (appearance, accessories, surrounding, identity, psychology). Each view contains multi-level attributes (\textbf{\textcolor[RGB]{13,93,135}{Level 1}}, \textbf{\textcolor[RGB]{196,96,26}{Level 2}}, \textbf{\textcolor[RGB]{99,138,102}{Level 3}}), comprising over 210 attributes and 700 \textbf{\textcolor[RGB]{134,106,163}{attribute values}} in total. \textbf{Right}: Q\&A of our Face-LLaVA finetuned using the FaceBench. Best viewed in color.}  
    \label{fig: overview}
    \vspace{-2pt}
\end{center}
}]

\footnoteONLYtext{*Corresponding author}
\begin{abstract}

Multimodal large language models (MLLMs) have demonstrated remarkable capabilities in various tasks. However, effectively evaluating these MLLMs on face perception remains largely unexplored. To address this gap, we introduce FaceBench, a dataset featuring hierarchical multi-view and multi-level attributes specifically designed to assess the comprehensive face perception abilities of MLLMs. Initially, we construct a hierarchical facial attribute structure, which encompasses five views with up to three levels of attributes, totaling over 210 attributes and 700 attribute values. Based on the structure, the proposed FaceBench consists of 49,919 visual question-answering (VQA) pairs for evaluation and 23,841 pairs for fine-tuning. Moreover, we further develop a robust face perception MLLM baseline, Face-LLaVA, by training with our proposed face VQA data. Extensive experiments on various mainstream MLLMs and Face-LLaVA are conducted to test their face perception ability, with results also compared against human performance. The results reveal that, the existing MLLMs are far from satisfactory in understanding the fine-grained facial attributes, while our Face-LLaVA significantly outperforms existing open-source models with a small amount of training data and is comparable to commercial ones like GPT-4o and Gemini. The dataset will be released at \href{https://github.com/CVI-SZU/FaceBench}{https://github.com/CVI-SZU/FaceBench}.

\end{abstract}
\section{Introduction}
\label{sec:intro}

In recent years, Multimodal Large Language Models (MLLMs) \cite{liu2024visual, zhu2023minigpt4, chung2024scaling, touvron2023llama, wu2023next, gao2023llama, zhu2023minigpt, awadalla2023openflamingo, achiam2023gpt, yin2024lamm} have demonstrated exceptional visual understanding capabilities across various domains, such as video analysis \cite{cheng2024videollama, luo2023valley}, autonomous driving \cite{xu2024drivegpt4}, embodied AI \cite{mu2024embodiedgpt}, remote sensing \cite{kuckreja2024geochat}, and medical applications \cite{li2024llava, moor2023med}. Inspired by the success of MLLMs, an increasing number of benchmarks have been developed to assess their diverse capabilities. For example, MMBench \cite{liu2025mmbench} and SEED-Bench \cite{li2024seed} evaluate the perception and reasoning abilities of MLLMs. Video-MME \cite{fu2024video} and MVBench \cite{li2024mvbench} assess the ability of MLLMs to handle sequential visual data. Additionally, GMAI-MMBench \cite{chen2024gmai} and OmniMedVQA \cite{hu2024omnimedvqa} examine their performance across various medical applications.

However, despite the growing amount of benchmarks being proposed, comprehensive evaluations specifically focused on face perception remain noticeably limited. We attribute this issue primarily to two key factors. \textbf{First}, human faces carry a wealth of information, necessitating MLLMs to analyze the face image from multiple perspectives. However, the current face visual question-answering (VQA) dataset, FABA-Instruct \cite{li2025facial}, focuses solely on emotion and facial action unit recognition. FairFace \cite{karkkainen2021fairface} only involves three facial attributes of age, gender, and race, which are used to highlight the view of identity. Both of them demonstrate significant view limitations. \textbf{Second}, for an MLLM to effectively mirror human perception, it must recognize not only superficial facial attributes but also deeper, more nuanced characteristics. Most existing face datasets, as indicated in Tab.~\ref{tab: comparsion}, such as CelebA-HQ \cite{karras2017progressive}, which comprises only 40 attributes with 80 binary values, and FFHQ-Text \cite{zhou2021generative}, featuring 162 attribute values across 13 groups, fall short in capturing the complex hierarchical structure and granularity of facial attributes perceived by humans. Consequently, these datasets fail to meet the comprehensive evaluation needs for MLLMs in face analysis, underscoring the necessity for more advanced benchmarks in this domain.

\begin{table*}
\centering
  \footnotesize
  \setlength{\tabcolsep}{2mm}{
  \begin{tabular}{cccccccccc}
  \toprule
  Datasets &\#Years &\#Images &\#Attributes &\#Values &Multi-views &Multi-levels &VQA &Manual  \\
  \midrule
  CelebA-HQ \cite{karras2017progressive} &2017 &30K  &40  &80   &\usym{2713}  & \usym{2717}  & \usym{2717} & \usym{2713} \\
  MM-CelebA-HQ \cite{xia2021tedigan}     &2021 &30K  &38  &76   &\usym{2713}  & \usym{2717}  & \usym{2717} & \usym{2713} \\
  CelebA-Dialog \cite{jiang2021talk}     &2021 &30K  &5   &--   &\usym{2713}  & \usym{2717}  & \usym{2713} & \usym{2713}  \\
  FFHQ-Text \cite{zhou2021generative}    &2021 &760  &13  &162  &\usym{2713}  & \usym{2717}  & \usym{2717} & \usym{2713} \\
  FairFace \cite{karkkainen2021fairface} &2021 &108K &3   &18   &\usym{2717}   & \usym{2717}  & \usym{2717} & \usym{2713} \\
  
  FABA-Instruct \cite{li2025facial}      &2024 &20K  &2   &26   &\usym{2717}   & \usym{2717}  & \usym{2713} & \usym{2717} \\
  FaceCaption-15M \cite{dai202415m}      &2024 &15M  &40  &--   &\usym{2713}   & \usym{2717} & \usym{2717} & \usym{2717} \\
  FLIP-80M \cite{li2024flip}             &2024 &80M  &--  &--   &\usym{2713}   & \usym{2717} & \usym{2717} & \usym{2717} \\
  \midrule
  Our FaceBench                          &2025 &16K  &211 &701  &\usym{2713}   & \usym{2713}   & \usym{2713}   & \usym{2713} \\
  \bottomrule
  \end{tabular}}
  \caption{Comparison of the number of images, facial attributes, and attribute values in different face datasets.}
  \label{tab: comparsion}
  \vspace{-6pt}
\end{table*}

To address these challenges, we introduce FaceBench, a multi-view, multi-level facial attribute benchmark. As illustrated in Fig.~\ref{fig: overview}, FaceBench is designed to comprehensively assess the face perception abilities of MLLMs. We begin by meticulously designing a hierarchical facial attribute structure, covering five distinct views: Appearance, Accessories, Surrounding, Psychology, and Identity. This structure is organized into three levels: Level 1 features coarse-grained attributes such as eyes and ears, Level 2 delves into components of these attributes like pupils and earlobes, and Level 3 offers a finer classification including size, type, color, and shape. Utilizing this framework, we construct a detailed face VQA dataset containing 49,919 pairs for evaluation and 23,841 pairs for training MLLMs. Additionally, we enhance the existing LLaVA \cite{liu2024visual} model by fine-tuning it with our FaceBench training set, thereby creating our specialized version, Face-LLaVA, to boost its face perception capabilities. Finally, we evaluate the performance of 12 MLLMs, including our Face-LLaVA, using the FaceBench dataset to demonstrate its effectiveness for face perception analysis.

The main contributions of this paper are summarized as follows:
\begin{itemize}
    \item We systematically develop a face perception structure, which comprehensively defines multi-views and multi-level attributes of the human face. This structure establishes a solid foundation for future study of face perception and generation tasks.
    \item Guided by our hierarchical facial attribute framework, we introduce FaceBench, a dataset tailored to the facial domain with multi-views and multi-level attributes. It includes 49,919 testing VQA pairs and 23,841 training pairs for instruction-based learning. To our knowledge, FaceBench is the most extensive face dataset available, offering unprecedented comprehensiveness in both the range of views and the depth of attributes.
    \item We conduct a thorough evaluation of face perception capabilities across 12 mainstream MLLMs, including an LLaVA model fine-tuned with our dataset. These extensive experiments demonstrate FaceBench's effectiveness for benchmarking MLLM performance in face understanding and enhancing their face perception abilities through fine-tuning.
\end{itemize}
\section{Related Work}
\label{sec:related_work}

\subsection{MLLMs and Facial MLLMs}
With the recent emergence of Large Language Models (LLMs) such as GPT \cite{ouyang2022training} and LLaMA \cite{touvron2023llama}, multimodal large language models (MLLMs) have begun leveraging LLM knowledge by aligning visual features with textual space to generate diverse text outputs. For example, Flamingo \cite{alayrac2022flamingo} incorporates visual features into the textual space by introducing cross-attention layers into LLMs. Similarly, BLIP-2 \cite{li2023blip} connects a pre-trained visual encoder with an LLM through an innovative Q-former, and InstructBLIP further enhances performance with instruction-following data. Building on these successes, LLaVA \cite{liu2024visual} constructs 665K instruction-following examples to improve training outcomes and achieve notable performance. Additionally, MiniGPT-4 \cite{zhu2023minigpt4}, Qwen-VL \cite{bai2023qwen}, mPLUG-Owl \cite{ye2023mplug}, and LLaMA-Adapter-v2 \cite{gao2023llama} have made significant contributions to the progress of MLLMs.

Inspired by the success of general-domain MLLMs, researchers have begun developing MLLMs specifically tailored for facial analysis. For instance, the Face Forgery Analysis Assistant (FFAA) \cite{huang2024ffaa} enhances LLaVA through fine-tuning to provide user-friendly, explainable results for open-world face forgery detection. Additionally, models such as EmoLA \cite{li2025facial}, EMO-LLaMA \cite{xing2024emo}, and Emotion-LLaMA \cite{cheng2024emotion} integrate facial prior knowledge into MLLMs to facilitate facial affective behavior analysis, including facial emotion and Action Unit (AU) recognition. Despite increasing interest in this field, there is still a lack of comprehensive evaluation of MLLMs specifically in the facial attribute perception.

\subsection{Facial VQA Datasets}
With the rapid development of facial MLLMs, several recent works have introduced VQA datasets for evaluating MLLM's performance on facial tasks. For example, OW-FFA-Bench \cite{huang2024ffaa} compiles a range of real and forged face images and uses GPT-4o to create the FFA-VQA dataset, which includes essential descriptions and forgery reasoning data. FABA-Bench \cite{li2025facial} introduces an instruction-following dataset, FABA-Instruct, focused on emotion and Action Unit (AU) recognition, containing 19K face images with 30K fine-grained annotations generated by GPT-4v. SHIELD \cite{shi2024shield} assesses MLLM's capabilities in face spoofing and forgery detection. EMO-LLaMA \cite{xing2024emo} aggregates publicly available facial expression recognition datasets and uses Gemini \cite{team2023gemini} to create a visual instruction dataset. MERR dataset \cite{cheng2024emotion} is designed for multimodal large models, supporting instruction tuning with diverse multimodal data, such as audio tone descriptions and visual objective descriptions. In contrast to prior works, we introduce FaceBench, a novel multimodal VQA dataset specifically designed for comprehensive face perception evaluation. We also establish a suite of evaluation standards to ensure the stability and accuracy of assessment results.
\section{FaceBench}
\label{sec:facebench}

\begin{figure*}[t]
  \vspace{-6pt}
  \centering
    \includegraphics[width=1.0\linewidth]{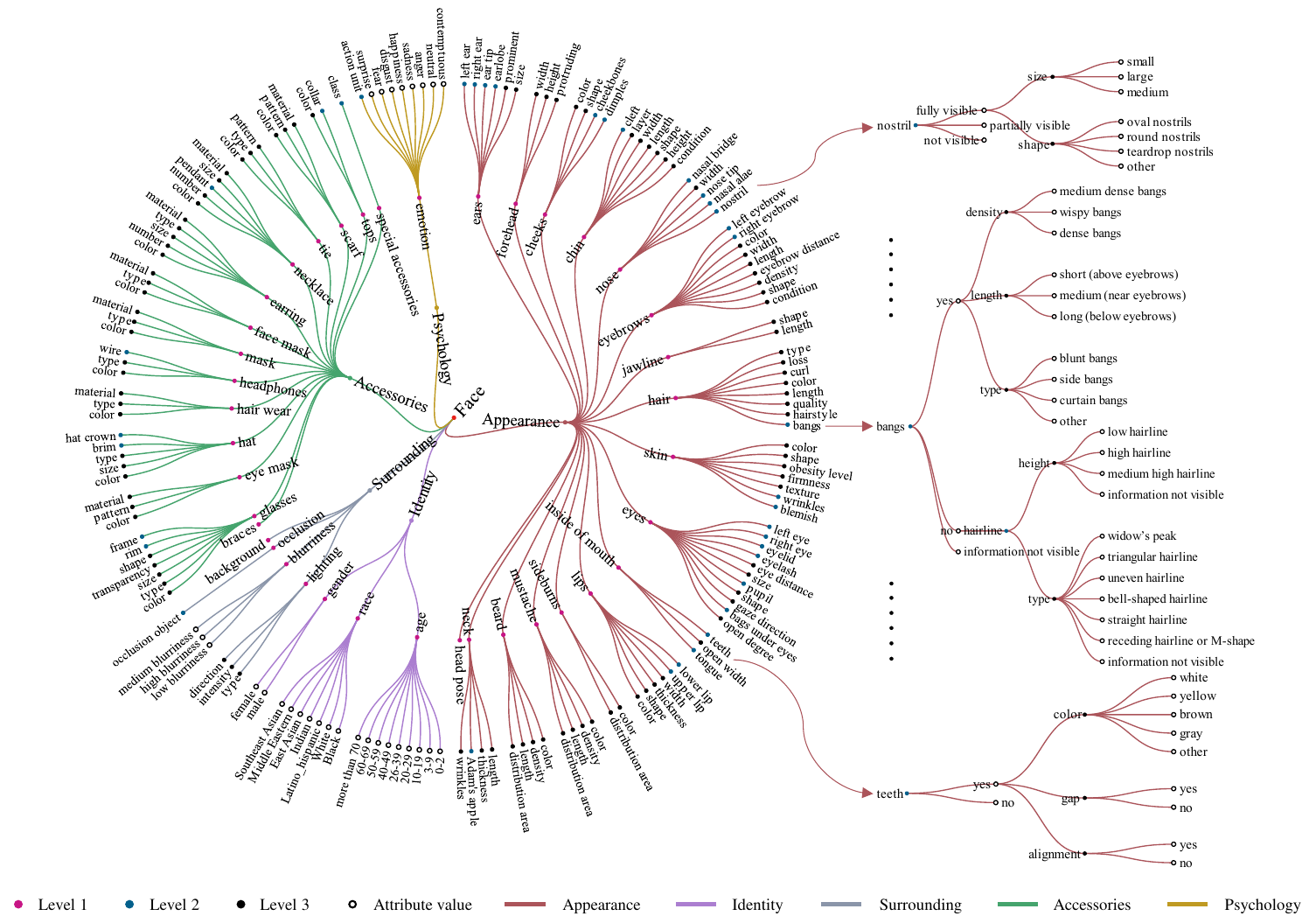}
    \caption{Hierarchical organization of facial attributes. We categorize facial attributes into Appearance, Identity, Surrounding, Accessories, and Psychology, illustrating their hierarchical structure across three levels. Best viewed in color.}
   \label{fig: hierarchical facial attribute structure}
   \vspace{-6pt}
\end{figure*}

\subsection{Hierarchical Facial Attribute Structure}
To enable a comprehensive evaluation of perceptual abilities in the facial domain, we first introduce a hierarchical structure of multi-view, multi-level facial attributes, as shown in Fig.~\ref{fig: hierarchical facial attribute structure}. This structure serves as the foundation of our FaceBench, organizing facial attributes into a systematic framework that captures diverse aspects and levels of detail, supporting in-depth analysis across multiple views.

\noindent\textbf{Multi-View Attributes.} 
Our hierarchical attribute structure defines five essential views for a comprehensive analysis of facial attributes within the dataset. \textbf{Appearance} captures fundamental biological features that shape facial structure, influenced by genetic and environmental factors, providing essential details for characterizing faces. \textbf{Psychology} includes attributes related to facial expressions, emotions, and social interactions, allowing for a nuanced examination of affective behaviors and psychological states. \textbf{Identity} conveys social meanings expressed through facial features, such as social identity, cultural background, and social roles, capturing the diversity of societal contexts within the dataset. \textbf{Accessories} cover items like jewelry, glasses, and other fashion elements that modify appearance, offering insights into personal expression and cultural significance. \textbf{Surrounding  } records the environmental context of the face, providing background details that enhance interpretive depth. As a result, these five views collectively support a detailed and multidimensional analysis of facial attributes in the dataset.

\noindent\textbf{Multi-level Attributes.} 
We further categorize facial attributes into multiple levels, mirroring the nuanced way humans perceive and interpret facial details. These levels enable a layered approach to analyzing facial attributes within the dataset.
\textbf{Level 1} encompasses fundamental facial features, such as eyes, ears, hair, age, emotion, and accessories like hats or masks, providing a baseline for recognizing primary attributes in face images. \textbf{Level 2} introduces finer subdivisions of Level 1 attributes, identifying detailed components—for example, the pupil within the eye, or the earlobe within the ear—enabling more granular analysis. \textbf{Level 3} further expands on Levels 1 and 2 by analyzing attributes across dimensions such as size, color, shape, and type, allowing for detailed characterization. Lastly, \textbf{Attribute Values} specify categorical variations within attributes, such as s-shaped, hard-angled, or straight for eyebrow shapes. This hierarchical structure supports a robust and scalable framework for analyzing facial attributes, facilitating a deeper exploration of face characteristics within the dataset.

\noindent\textbf{Attribute Statistics.} 
Following our hierarchical attribute structure, we systematically collect a comprehensive set of attributes and values from referenced sources~\cite{wu2023logical} and online resources\footnote{ \scriptsize https://descriptionary.wordpress.com/give-your-characters-life/}\footnote{\scriptsize https://www.faceplusplus.com.cn/facial-features/}\footnote{\scriptsize https://www.heatwaveworcester.co.uk/tanning-worcester/fitzpatrick-skin-type/}\footnote{\scriptsize https://www.fhiheat.com/blogs/journal/what-is-your-hair-type}\footnote{\scriptsize https://www.westlakedermatology.com/blog/common-types-of-facial-wrinkles/}. Tab.~\ref{tab: statistics of hierarchical facial attribute structure} provides an overview of each defined view, listing the number of attributes across different levels and their corresponding attribute values. Our dataset comprises 5 primary views, with 39 Level 1 attributes, 39 Level 2 attributes, and 133 Level 3 attributes, for a total of 211 attributes and 701 attribute values. The emotion and facial action unit labels are drawn from the RAF-DB \cite{li2017reliable} and RAF-AU \cite{yan2020raf} datasets, respectively, while identity labels are sourced from the FairFace \cite{karkkainen2021fairface} dataset. Detailed definitions for each hierarchical attribute level are included in the Appendix. This dataset thus provides a modular and scalable structure, enabling comprehensive, multi-dimensional analysis of facial images across diverse attribute categories.

\begin{table}
  \centering
    \footnotesize
  \begin{tabular}{ccccccc}
    \toprule
    Attributes & AP & AC & SU & PS & ID & Overall\\
    \midrule
    Level 1 &18 &13 &4 &1 &3 &39 \\
    Level 2 &29 &8 &1 &1 &/ &39 \\
    Level 3 &85 &45 &3 &/ &/ &133 \\
    \midrule
    Values  &430 &200 &19 &34 &18 &701 \\
    \midrule
    Templates &121 &61 &7 &2 &3 &194 \\
    \bottomrule
  \end{tabular}
  \caption{Summary of the statistics of facial attributes, attribute values, and VQA templates in our dataset. The five primary views—AP (Appearance), AC (Accessories), SU (Surrounding), PS (Psychology), and ID (Identity)—are organized across three hierarchical levels. The table presents the number of attributes at each level, the total number of attribute values, and VQA templates. '/' indicates None.}
  \label{tab: statistics of hierarchical facial attribute structure}
\end{table}

\begin{figure}[t]
  \centering
   \includegraphics[width=1.0\linewidth]{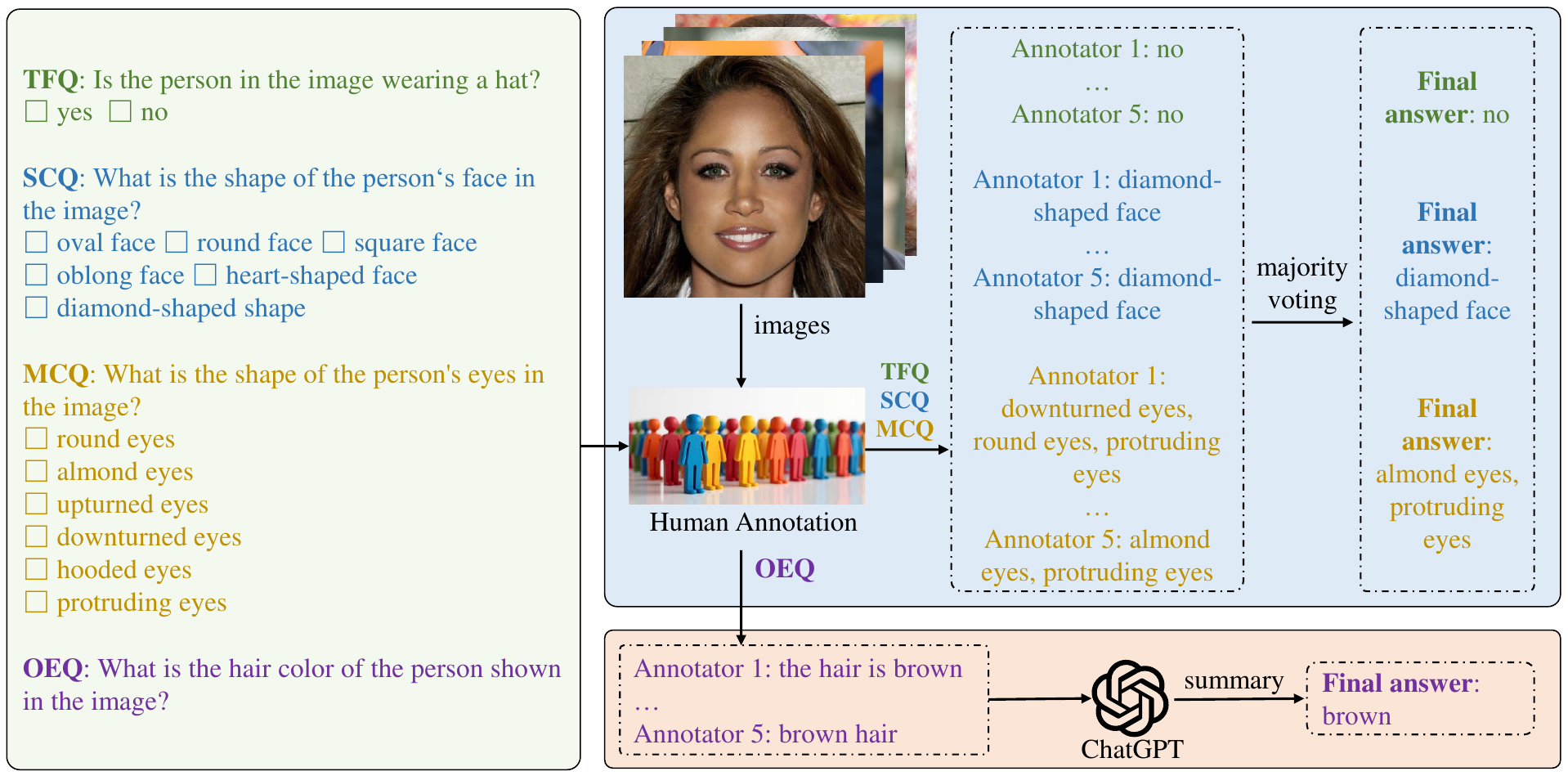}
   \caption{Question types and human annotation workflow for building our dataset. Best viewed in color.}
   \label{fig: system}
   \vspace{-6pt}
\end{figure}

\subsection{Dataset Collection and Annotation}
Building on our hierarchical facial attribute structure, we construct FaceBench—a facial Visual Question Answering (VQA) benchmark for detailed, multi-dimensional analysis of facial attributes. It includes diverse question-answer pairs reflecting various aspects of facial perception, developed through the following four key steps.

\noindent\textbf{Image Collection.} 
To comprehensively address the five attribute views in our hierarchical facial attribute structure—Identity, Psychology, Appearance, Accessories, and Surrounding—we compile a diverse image collection of 15,842 face images from multiple datasets. For the Identity view, we include the FairFace \cite{karkkainen2021fairface} test set with 10,954 images labeled by attributes such as race, gender, and age, supporting analysis of social identity contexts. Additionally, we randomly selected 200 images to enrich the training set of our FaceBench. Psychology attributes are covered using the RAF-DB \cite{li2017reliable} and RAF-AU \cite{yan2020raf} datasets. RAF-DB provides 3,068 images labeled across seven emotional categories for analyzing affective states, with 200 images randomly selected from the resting images to form the training set in our FaceBench. We randomly selected 200 images for the training set and 920 images for the test set in the RAF-AU dataset to examine facial muscle movements linked to psychological cues. For the Appearance, Accessories, and Surrounding views, we select high-resolution images from CelebA-HQ \cite{karras2017progressive} and FFHQ \cite{karras2019style}. CelebA-HQ, a refined subset of the CelebA \cite{liu2015deep} celebrity dataset, and FFHQ, sourced from Flickr, offer diverse representations of facial details, fashion accessories, and backgrounds. To ensure balanced representation, we randomly sample 90 images from CelebA-HQ and 210 from FFHQ, totaling 300 images, for manual annotation. We further split the 300 annotated images into development and test subsets. The development subset, containing 193 images, is intended for fine-tuning existing models, while the remaining 107 images form the FaceBench test set. This curated dataset comprehensively covers all five attribute views, enabling robust, multi-dimensional facial attribute analysis.

\begin{figure*}[t]
  \centering
   \includegraphics[width=1.0\linewidth]{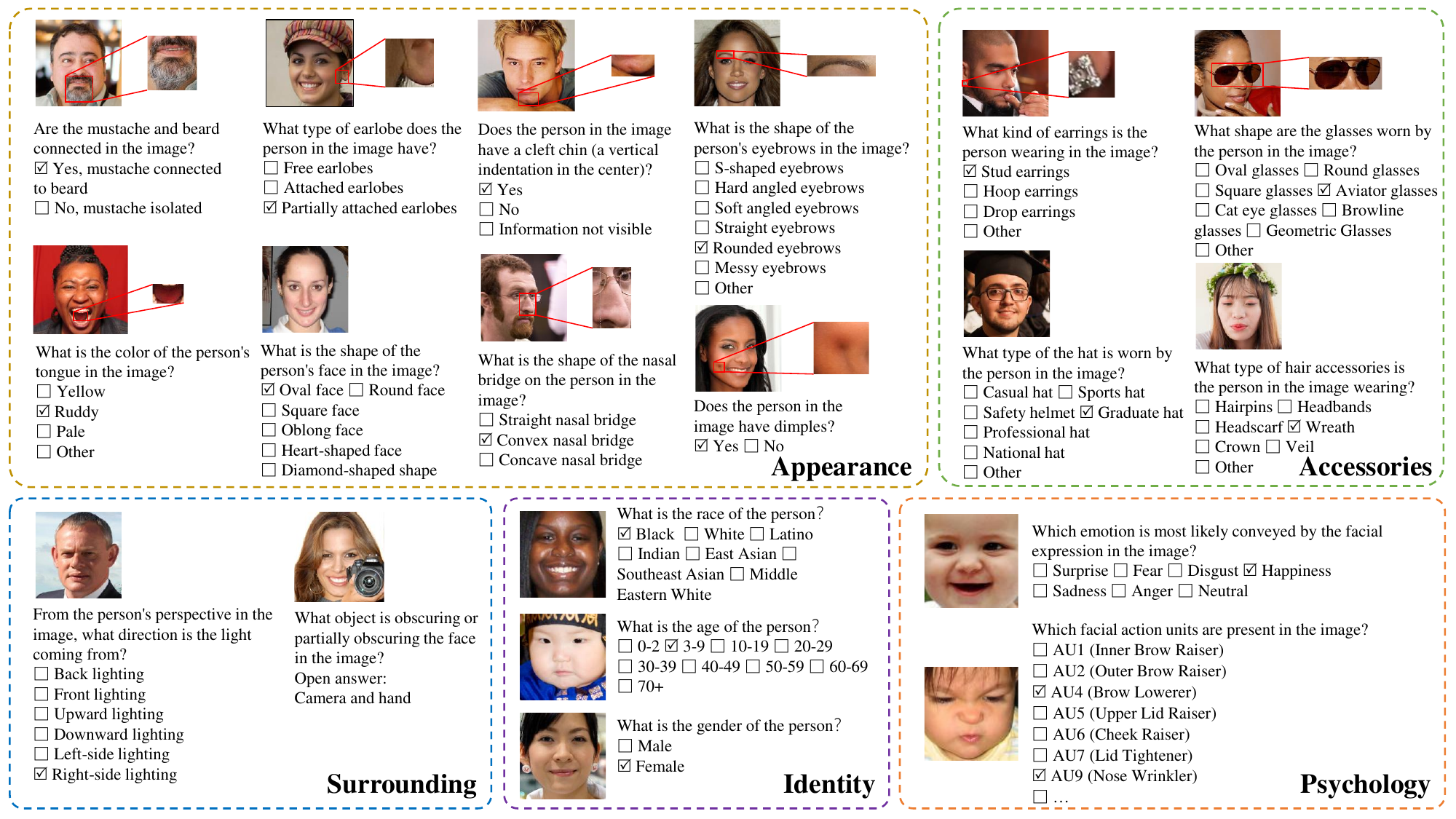}
   \caption{Samples from our FaceBench dataset. It displays a range of VQA pairs from our dataset aimed at evaluating the perception of facial attributes categorized into Appearance, Accessories, Surrounding, Identity, and Psychology.}
   \label{fig: example}
\end{figure*}

\begin{table*}
    \centering
    \footnotesize
    \setlength{\tabcolsep}{1.8mm}{
    \begin{tabular}{c|ccccc|ccccccc}
    \toprule
    \multirow{3}{*}{\#VQA} & \multicolumn{5}{c|}{Test Set} & \multicolumn{5}{c}{Training Set} \\
    \cmidrule(r){2-6} \cmidrule(r){7-11}
    & Appearance & Accessories & Surrounding & Psychology & Identity & Appearance & Accessories & Surrounding & Psychology & Identity \\
    \cmidrule(r){1-11}
    TFQ &2,905  &1,401  &107   &/       &/        &5,189    &2,490   &190     &/      &/ \\
    SCQ &6,799  &643    &428   & 3,068  & 32,862  &12,100   &859     &763     &200    &600 \\
    MCQ &166    &/      &/     & 920    &/        &257      &/       &/       &200    &/ \\
    OEQ &297    &207    &116   &/       &/        &519      &273     &201     &/      &/ \\
    \cmidrule(r){1-6} \cmidrule(r){7-11}
    Overall & \multicolumn{5}{c|}{49,919} &\multicolumn{5}{c}{23,841} \\
    \toprule
  \end{tabular}}
  \caption{Distribution of visual question-answer pairs for both the training and test splits in FaceBench. '/' indicates None.}
  \label{tab: statistics}
  \vspace{-6pt}
\end{table*}

\noindent\textbf{Question Templates.} 
To translate the hierarchical facial attribute structure into a Visual Question Answering (VQA) format, we create a diverse set of question templates that cover each attribute. As shown in Fig.~\ref{fig: system}, we design four question types—true/false (TFQ), single-choice (SCQ), multiple-choice (MCQ), and open-ended question (OEQ)—based on our hierarchical attribute structure. This ensures a varied and comprehensive set of QA pairs. For example, a TFQ might be, ``Is the person in the image wearing a hat? Options: Yes; No," while an OEQ could be, ``What is the hair color of the person shown in the image?" In total, we design 194 question templates across the five views, providing broad coverage of facial attributes and question formats. Tab.~\ref{tab: statistics of hierarchical facial attribute structure} details the template distribution by different views, supporting robust, multi-dimensional evaluation of model performance.

\noindent\textbf{Human Annotation.} 
To provide comprehensive, fine-grained attribute annotations for the 300 face images from CelebA-HQ and FFHQ, we develop a web-based system for manual annotation, and the detailed information is included in the Appendix. We recruit 200 college and graduate student volunteers, each with a basic understanding of facial attributes, to perform the annotations independently. During each annotation session, the system presents a randomly selected image along with a set of questions related to the six Level 1 attributes to minimize labeling errors. To enhance efficiency, in-depth attribute questions are dynamically displayed based on previous answers. For instance, if an annotator indicates that the person is not wearing a hat, further questions about hats are skipped. Each Level 1 attribute is annotated by five different volunteers to improve reliability. Fig.~\ref{fig: system} shows the human annotation workflow, and we determine the final ground-truth labels through majority voting for true/false (TFQ) and single-choice (SCQ) questions. For multiple-choice questions (MCQ), we set a threshold of $\alpha = 0.4$, where an option is accepted as correct if chosen by more than 40\% of annotators. For open-ended questions (OEQ), we use ChatGPT\footnote{https://openai.com/chatgpt} to synthesize a comprehensive answer from all responses. 

\noindent\textbf{Quality Control.} To ensure high-quality annotations, we set the following requirements: (1) Each annotator undergoes training before starting to ensure a correct understanding of each question. (2) Within the annotation system, we provide a face mask reference image (Fig.~\ref{fig: overview}) to clarify any challenging questions. (3) We also supply reference images for specific answer options to minimize ambiguity. (4) Additionally, we record the time taken for each response; answers given too quickly are flagged as potential errors and discarded. Finally, each annotation is cross-checked by two senior annotators to enhance accuracy and reduce errors caused by human bias.

\subsection{Dataset Statistics and Analysis} 
We gather a total of 15,842 face images from various sources and meticulously organize the collected images into a test set of 15,049 images and a training set of 793 images. Each set includes four question types and five views, with 49,919 VQA pairs in the test set and 23,841 VQA pairs in the training set. Overall, FaceBench consists of 73,760 VQA pairs across five views and three levels of attributes, designed to assess the face perception capability of MLLMs. The detailed VQA counts for the test and training sets are listed in Tab.~\ref{tab: statistics}, and Fig.~\ref{fig: example} illustrates representative VQA pairs from different views. More detailed statistics and examples of FaceBench are available in the Appendix.
\section{Experiments}
\label{sec:experiments}

\begin{table*}
  \centering
  \footnotesize
  \begin{tabular}{cccccccccc}
    \toprule
    Models &Appearance &Accessories &Surrounding &Psychology &Identity & Overall\\
    \midrule
    Qwen-VL-Chat-7B \cite{bai2023qwen} &48.13 &57.79 &42.04 &43.07 & 49.39 &48.27  \\ 
    InstructBLIP-7B \cite{dai2023instructblip} &44.33 &58.26 &52.96 &48.64 &62.86 &51.62 \\
    DeepSeek-VL-7B \cite{lu2024deepseek} &49.24 &65.92 &\textbf{{\color{red}61.77}} &41.77 &65.66 &56.09  \\
    Qwen2-VL-7B \cite{wang2024qwen2} &53.89 &70.73 &58.35 &37.81 &65.07 &57.19 \\
    Molmo-7B\footnote[7]  &55.41 &65.19 &46.18 &39.08 &68.63 &54.04 \\
    InternVL2-8B \cite{chen2024internvl}  &51.59 &62.42 &59.27 &57.43 &63.71 &57.69 \\
    MiniCPM-V-8B \cite{yao2024minicpm}  &53.42 &\textbf{{\color{red}70.95}} &58.75 &59.57 &66.76 &60.67 \\
    LLaMA-3.2-11B\footnote[8]  &47.74 &58.48 &26.52 &44.66 &60.97 &45.87 \\
    LLaVA1.5-13B \cite{liu2024visual}  &51.22 &58.45 &44.11 &58.47 &68.10 &53.66 \\
    \rowcolor{mygray}
    Face-LLaVA-13B &\textbf{{\color{red}60.29}} &61.62 &57.89 &\textbf{{\color{red}61.85}} &\textbf{{\color{red}71.64}} &\textbf{{\color{red}61.16}}\\
    \midrule
    GPT-4o \cite{achiam2023gpt} &\textbf{{\color{blue}60.69}} &\textbf{{\color{blue}68.23}} &53.63 &\textbf{{\color{blue}74.94}} &63.53 &\textbf{{\color{blue}63.21}} \\
    Gemini-1.5-Pro \cite{team2024gemini} &58.74 &66.55 &\textbf{{\color{blue}57.20}} &68.98 &\textbf{{\color{blue}71.11}} &62.72 \\
    \midrule
    Human &\textbf{67.49} &\textbf{75.96} &\textbf{63.04} &-- &-- &\textbf{67.38}\\
    \bottomrule
  \end{tabular}
   \caption{Comparative performance of various multimodal large language models across five facial attribute views. The {\color{red}red} text indicates the best results for open-source models, and the {\color{blue}blue} indicates the best results for commercial models.}
  \label{tab: multi-views}
  \vspace{-6pt}
\end{table*}

\subsection{Experimental Setup}

\textbf{Metrics.}
To effectively evaluate MLLMs given the diversity of question types, we adopt distinct evaluation metrics tailored to each format. For TFQ and SCQ, we measure the accuracy of the selected options. For MCQ, we assess model performance using the macro F1 score. For OEQ, we use the ROUGE-L score to evaluate the accuracy of the responses. Some MLLMs may generate lengthy responses instead of directly selecting an option for TFQ, SCQ, and MCQ questions. In such cases, we use ChatGPT to interpret the response and match it to the closest option. Finally, to provide an overall accuracy measure, we calculate the average score across the metrics for all problems.

\noindent\textbf{Baselines.} We evaluate a total of nine advanced MLLMs: Qwen-VL-Chat-7B \cite{bai2023qwen}, InstructBLIP-7B \cite{dai2023instructblip}, DeepSeek-VL-7B \cite{lu2024deepseek}, Qwen2-VL-7B \cite{wang2024qwen2}, Molmo-7B\footnote{https://huggingface.co/allenai/Molmo-7B-D-0924}, InternVL2-8B \cite{chen2024internvl}, MiniCPM-V-8B \cite{yao2024minicpm}, LLaMA-3.2-11B\footnote{https://huggingface.co/meta-llama/Llama-3.2-11B-Vision}, and LLaVA1.5-13B \cite{liu2024visual}. Additionally, we assess two commercial models: GPT-4o \cite{achiam2023gpt}, Gemini-1.5-Pro \cite{team2024gemini}, using their official configurations.

\noindent\textbf{Face-LLaVA.} To demonstrate the effectiveness of our manually annotated data and the value of our benchmark, we fine-tune LLaVA1.5-13B \cite{liu2024visual} using our training set, consisting of 23,841 VQA pairs. Details on the fine-tuning method and parameters are provided in the Appendix.

\subsection{Evaluation Results and Analysis}

\rowcolors{2}{gray!15}{white}
\begin{table*}
  \scriptsize
  \centering
  \setlength{\tabcolsep}{1.0mm}{
  \begin{tabular}{c|cccccccc|cc|cc}
    \toprule
    Attributes &Qwen-VL-Chat &InstructBLIP &DeepSeek-VL &Qwen2-VL &Molmo & InternVL2 & MiniCPM-V & LLaMA & LLaVA & Face-LLaVA & GPT-4o & Gemini\\
    \hline
    earring   &53.67 &55.11 &64.68 &70.16 &58.52 &60.53 &{\color{red}75.75} &60.89 &52.77 &48.43 &70.30 &{\color{blue}70.39}  \\
    eye mask  &65.44 &66.05 &66.05 &99.39 &82.72 &82.72 &{\color{red}99.70} &65.75 &98.77 &99.69 &{\color{blue}99.38} &49.70  \\
    face mask &99.08 &99.08 &99.08 &99.08 &{\color{red}100.00} &97.24 &{\color{red}100.00} &99.08 &97.24 &{\color{red}100.00} &{\color{blue}100.00} &{\color{blue}100.00}  \\
    glasses   &70.41 &66.74 &80.22 &80.05 &70.96 &70.39 &{\color{red}83.32} &73.77 &70.90 &73.56 &{\color{blue}81.64} &74.48  \\ 
    hair wear   &44.93 &58.90 &55.34 &66.56 &58.80 &57.53 &{\color{red}67.94} &51.15 &45.27 &60.15 &{\color{blue}63.46} &59.39  \\
    hat   &67.44 &68.07 &73.85 &74.60 &72.23 &71.60 &{\color{red}77.63} &67.80 &68.05 &73.42 &75.88 &{\color{blue}76.49}  \\ 
    headphones   &82.77 &91.70 &75.30 &{\color{red}95.39} &90.16 &83.87 &92.01 &88.31 &87.91 &87.31 &85.15 &{\color{blue}89.63}  \\
    mask   &96.31 &{\color{red}99.08} &98.16 &{\color{red}99.08} &{\color{red}99.08} &93.55 &{\color{red}99.08} &98.16 &94.47 &{\color{red}99.08} &99.07 &{\color{blue}99.10} \\ 
    necklace   &53.30 &51.39 &59.09 &{\color{red}61.28} &55.42 &59.58 &57.16 &50.38 &54.84 &50.05 &46.67 &{\color{blue}57.82}  \\
    scarf   &56.74 &54.30 &62.03 &34.23 &52.15 &49.66 &{\color{red}63.26} &39.50 &50.59 &57.81 &49.89 &{\color{blue}56.83}  \\ 
    special accessories   &14.33 &{\color{red}41.24} &24.68 &27.45 &21.06 &25.25 &24.29 &10.45 &12.76 &33.37 &17.29 &{\color{blue}21.14} \\
    tie   &65.18 &64.33 &{\color{red}66.17} &65.25 &{\color{red}66.17} &59.68 &{\color{red}66.17} &65.25 &56.96 &65.50 &{\color{blue}53.27} &48.17  \\ 
    top   &50.43 &27.05 &57.43 &{\color{red}68.53} &63.56 &52.60 &58.26 &48.02 &54.74 &45.13 &{\color{blue}67.96} &59.10  \\
    bread   &52.71 &54.72 &65.64 &{\color{red}74.63} &62.98 &58.99 &61.19 &59.53 &67.93 &66.93 &{\color{blue}67.75} &66.78  \\ 
    cheeks   &60.31 &48.81 &41.61 &52.51 &56.73 &50.42 &67.33 &61.83 &47.49 &{\color{red}75.59} &68.12 &{\color{blue}68.69}  \\
    chin   &44.11 &50.74 &53.45 &43.01 &54.14 &49.79 &63.67 &58.87 &41.59 &{\color{red}66.34} &{\color{blue}70.39} &69.56  \\ 
    ears   &49.35 &37.33 &40.92 &{\color{red}60.14} &55.62 &46.68 &59.55 &44.40 &43.95 &59.07 &{\color{blue}56.95} &35.84  \\
    eyebrows   &61.83 &25.42 &43.46 &63.39 &69.57 &44.26 &67.04 &56.95 &61.82 &{\color{red}77.32} &70.15 &{\color{blue}72.90}  \\ 
    eyes   &58.13 &52.12 &59.32 &66.53 &70.23 &63.80 &65.19 &62.71 &62.42 &{\color{red}71.56} &74.58 &{\color{blue}76.10}  \\
    forehead   &39.56 &47.54 &38.63 &58.24 &41.96 &47.60 &58.24 &43.54 &42.57 &{\color{red}62.89} &48.13 &{\color{blue}63.06}  \\ 
    hair   &57.54 &56.01 &62.35 &63.77 &{\color{red}63.83} &60.21 &63.30 &53.79 &58.84 &60.14 &{\color{blue}68.67} &60.35  \\
    head pose   &48.13 &27.61 &75.70 &51.09 &72.23 &56.91 &67.69 &28.19 &72.87 &{\color{red}75.76} &{\color{blue}79.58} &71.41  \\ 
    inside of mouth   &52.20 &49.46 &66.61 &67.78 &66.04 &63.21 &{\color{red}70.73} &58.98 &60.96 &67.62 &67.39 &{\color{blue}68.44}  \\
    jawline   &52.39 &39.72 &55.00 &39.26 &{\color{red}61.00} &53.18 &51.51 &60.16 &54.92 &54.78 &55.15 &{\color{blue}61.87} \\ 
    lips   &45.38 &22.12 &45.55 &48.80 &52.30 &51.10 &58.15 &49.49 &48.94 &{\color{red}67.33} &{\color{blue}59.82} &52.49   \\
    mustache   &53.44 &51.39 &61.78 &{\color{red}67.91} &61.82 &56.04 &57.92 &47.46 &63.04 &57.15 &{\color{blue}71.82} &61.87  \\ 
    neck   &35.23 &38.87 &28.67 &35.82 &51.63 &47.51 &38.51 &49.95 &43.96 &{\color{red}52.32} &38.73 &{\color{blue}40.98}   \\
    nose   &68.30 &17.99 &63.71 &68.08 &69.80 &57.01 &63.01 &{\color{red}70.14} &67.48 &68.08 &{\color{blue}73.76} &72.03  \\ 
    sideburns   &49.36 &42.64 &48.54 &54.48 &{\color{red}56.29} &50.54 &45.96 &37.66 &50.42 &42.31 &{\color{blue}55.40} &54.12  \\
    skin   &57.63 &52.73 &59.90 &{\color{red}64.78} &60.18 &60.70 &60.82 &51.85 &54.41 &62.81 &67.83 &{\color{blue}69.04}   \\ 
    action unit  &49.71 &49.14 &27.97 &20.71 &20.23 &{\color{red}51.88} &50.15 &39.54 &49.89 &49.91 &{\color{blue}67.68} &64.41   \\
    emotion   &36.42 &48.04 &55.76 &55.07 &57.88 &63.01 &68.79 &49.84 &66.89 &{\color{red}73.60} &{\color{blue}82.69} &74.06   \\ 
    age   &27.23 &33.37 &45.66 &39.79 &43.04 &39.78 &48.05 &35.00 &45.76 &{\color{red}53.38} &{\color{blue}58.89} &52.00  \\
    gender   &89.54 &96.22 &92.97 &93.46 &95.36 &92.56 &92.65 &91.82 &96.52 &{\color{red}96.60} &70.26 &{\color{blue}94.43}  \\ 
    race   &31.87 &59.50 &58.64 &61.87 &{\color{red}67.70} &59.04 &59.85 &56.04 &62.33 &65.11 &60.01 &{\color{blue}66.65}  \\
    background   &3.24 &4.77 &{\color{red}18.60} &17.86 &4.85 &13.01 &11.78 &17.34 &6.42 &2.92 &2.40 &{\color{blue}4.39}  \\ 
    blurriness   &90.43 &{\color{red}96.17} &{\color{red}96.17} &95.22 &83.25 &90.91 &{\color{red}96.17} &31.92 &16.19 &{\color{red}96.17} &91.56 &{\color{blue}92.94}  \\
    lighting   &54.39 &48.65 &65.23 &63.32 &41.31 &{\color{red}73.05} &57.89 &61.56 &42.98 &62.84 &65.73 &{\color{blue}70.89}  \\ 
    occlusion   &45.27 &57.26 &67.41 &70.09 &67.20 &65.01 &{\color{red}74.41} &17.52 &70.62 &73.96 &{\color{blue}74.44} &49.52  \\ 
    \bottomrule
  \end{tabular}}
  \caption{Comparative performance of various MLLMs and our Face-LLaVA across Level 1 facial attributes in our dataset. The {\color{red}red} text indicates the best results for open-source models, and the {\color{blue}blue} indicates the best results for commercial models.}
  \label{tab:level1-attribute}
  \vspace{-6pt}
\end{table*}
\rowcolors{2}{}{}

\noindent\textbf{Results on Different Views.}
Tab.~\ref{tab: multi-views} displays the results comparing the face perception abilities of various MLLMs in individual views. In the Appearance category, GPT-4o outperforms all other MLLMs with a score of 60.69\%, highlighting that accurately capturing the overall appearance of a face remains challenging for current models. For Accessories, scores range significantly from approximately 57\% to 70\%, indicating that MLLMs are generally better at recognizing external facial attributes than intrinsic features. In the Surrounding category, model performance varies widely between 26\% and 61\%, underscoring substantial disparities in how models perceive environmental context. Psychologically, GPT-4o markedly outperforms other models, particularly open-source MLLMs, in tasks involving emotion and facial action unit analysis. In the Identity category, Gemini-1.5-Pro shows the highest performance among existing MLLMs, with LLaVA1.5-13B also yielding comparable results. Overall, the varied performance of current MLLMs on our dataset highlights the potential for improvement and underscores the challenges that even state-of-the-art commercial models face in this domain.

\noindent\textbf{Results on Different Attributes.}
Tab.~\ref{tab:level1-attribute} showcases a comprehensive evaluation of various MLLMs across 39 Level 1 facial attributes. The assessment highlights a wide range of performance outcomes for each model across specific attributes such as earrings, face masks, and glasses, with accuracy scores extending from as low as 22.12\% for lips recognition by InstructBLIP to as high as 100\% for face mask recognition by several models including Molmo and MiniCPM-V. Notably, commercial models like Gemini and GPT-4o exhibit robust across-the-board performances, particularly excelling in recognizing glasses, headphones, head pose, and masks. The data indicates that while some models demonstrate strong capabilities in perceiving external accessories, performance varies more significantly with intrinsic facial features such as eyebrows, nose, and lips. This variability underscores the ongoing challenges in achieving consistent high-level accuracy across more nuanced facial attributes, highlighting specific areas where future model training can be enhanced to improve overall performance.

\noindent\textbf{Results of Face-LLaVA.}
Tab.~\ref{tab: multi-views} shows that our fine-tuned model, Face-LLaVA, outperforms nine open-source MLLMs in key perception views, nearly matching GPT-4o in Appearance with a score of 60.29\%, and achieving the highest score of 71.64\% in Identity. This performance is due to our use of high-quality, hierarchical facial attribute data which enhances facial feature recognition. With strong results in Psychology at 61.85\%, Face-LLaVA demonstrates robust capabilities, though it still trails behind human-level performance in Accessories and Surrounding. Due to baseline limitations and a limited number of images, Face-LLaVA's performance may not be the highest in certain attributes, such as recognizing earrings and background details. Despite this, Face-LLaVA delivers strong overall performance, as shown in Tab.~\ref{tab:level1-attribute}. These findings emphasize the impact of our instruction-following dataset and highlight the need for further advancements in facial data and model architectures to improve face perception.

\noindent\textbf{Human Evaluation.} 
To reduce the cost of manual evaluation, we enlist 11 evaluators, separate from our annotators, to assess 10 randomly selected images from the test set within the views of Appearance, Accessories, and Surrounding. Then, we extracted 1,239 VQA pairs for manual evaluation and analysis. Tab.~\ref{tab: multi-views} shows the average score of their answers. It reveals that GPT-4o currently lags behind human evaluators by approximately 7.98\% across these three views, underscoring the challenges of our FaceBench dataset and the limitations of existing MLLMs in processing hierarchical facial attributes effectively. It is worth noting that due to the low image resolution from the RAF-DB, RAF-AU, and FairFace datasets, we did not perform human evaluation on the views of Psychology and Identity.
\section{Conclusion}
\label{sec:conclusion}
This paper introduces FaceBench, a comprehensive benchmark for assessing MLLMs in face perception. It leverages a multi-view and multi-level attribute analysis, supported by 200 annotators to ensure data quality and minimize bias. Our FaceBench dataset contains 15,842 images and 73,760 question-answer items, covering five views, over 210 attributes, and more than 700 attribute values. Additionally, We develop Face-LLaVA, a robust face perception MLLM, utilizing this dataset. Our evaluations of 11 widely-used MLLMs and Face-LLaVA demonstrate significant performance gaps compared to human face perception, underscoring the effectiveness of FaceBench in advancing multi-view multi-level facial attribute analysis technologies.

\section*{Acknowledgement}
This work was supported by the National Key R$\&$D Program of China (No. 2024YFF0618403), National Natural Science Foundation of China under Grant 82261138629 and 62206180, Guangdong Basic and Applied Basic Research Foundation under Grant 2023A1515010688, Guangdong Provincial Key Laboratory under Grant 2023B1212060076, and XJTLU Research Development Funds under Grant RDF-23-01-053.

{
    \small
    \bibliographystyle{ieeenat_fullname}
    \bibliography{main}
}

\end{document}